\documentclass[%
  aip,
  jcp,
  amsmath,amssymb,
  reprint,%
  floatfix,%
]{revtex4-1}

\usepackage{graphicx}%
\usepackage{dcolumn}%
\usepackage{bm}%

\usepackage[utf8]{inputenc}
\usepackage[T1]{fontenc}
\usepackage{mathptmx}
\usepackage{etoolbox}

\usepackage{amsmath}
\usepackage{amsfonts}
\usepackage{mathtools}
\usepackage{physics}
\usepackage{bbm}
\usepackage{amssymb}
\usepackage{url}
\usepackage{aligned-overset}
\usepackage{subcaption}
\usepackage[ruled,vlined]{algorithm2e}

\usepackage[hidelinks]{hyperref}

\newtheorem{definition}{Definition} %

\newcommand{\vecr}{\mathbf{r}}

\newcommand{\vp}{\mathbf{p}}

\renewcommand{\vu}{\mathbf{v}}

\newcommand{\qreset}{q_{\text{reset}}}

\DeclareMathOperator*{\argmin}{arg\,min}

\makeatletter
\def\@email#1#2{%
  \endgroup
  \patchcmd{\titleblock@produce}
  {\frontmatter@RRAPformat}
  {\frontmatter@RRAPformat{\produce@RRAP{*#1\href{mailto:#2}{#2}}}\frontmatter@RRAPformat}
  {}{}
}%
\makeatother
\begin{document}

\preprint{AIP/123-QED}

\title[Surrogate Functionals]{Surrogate Functionals for Machine-Learned Orbital-Free Density Functional Theory}

\author{Roman Remme} %
\author{Fred A. Hamprecht}%
\affiliation{
$^1$ Interdisciplinary Center for Scientific
Computing (IWR), Heidelberg University %
}%

\date{\today}%

\begin{abstract}
We introduce \textit{surrogate functionals}: machine-learned energy functionals for orbital-free density functional theory (OF-DFT) which are defined not by universal fidelity to a physical reference, but merely by the requirement that density optimization with a fixed procedure yields the true ground-state density. Helpfully, training surrogate functionals requires only ground-state densities, no energies or gradients away from the ground state. We here propose a gradient-descent-improvement loss that guarantees exponential convergence of the density to the ground state, and combine it with an adaptive sampling scheme that concentrates learning around the optimization trajectories actually visited during inference. On the QM9 and QMugs benchmarks, surrogate functionals achieve density errors competitive with or improving upon the state of the art for fully supervised machine-learned OF-DFT, while eliminating the need for the $O(N^3)$ orthononormalization step required by prior work, yielding improved runtime scaling for larger systems.
\end{abstract}

\maketitle

\makeatletter
\def\blfootnote{\gdef\@thefnmark{}\@footnotetext}
\makeatother

\maketitle
\blfootnote{This preprint has been submitted to The Journal of Chemical Physics.}

\section{\label{sec:level1}Introduction}

Kohn--Sham density functional theory (KS-DFT)\cite{kohn1965self} is one of the central tools in electronic-structure calculations, but its computational cost impedes its use for large systems and, when studying dynamics, long time scales.
Orbital-free DFT (OF-DFT) raises hopes of more favorable scaling by eschewing orbitals and instead minimizing an electronic energy functional with respect to the electron density.
In practice, however, OF-DFT hinges on the availability of accurate approximations---in particular for the kinetic energy\cite{thomas1927calculation, fermi1928statistische, von1935theorie}---and on robust, efficient density optimization.

Machine learning has recently emerged as an enabling technology to construct OF-DFT functionals\cite{chen2026machine, snyder2013orbital,yao2016kinetic,seino2018semi-local,fujinami2020orbital-free,meyer2020machine,kineticnet,M-OFDFT,remme2025stable}.
Much of the existing work aims at approximating a concrete, physical energy functional as faithfully as possible, ideally across broad chemical space and for arbitrary input densities.
In this contribution, we take a step back and lift this constraint in favor of a more solution-focused viewpoint: for many applications, the key requirement is not that a learned functional matches the true functional \emph{everywhere}, but that it enables reliable OF density optimization while realizing the promised scaling improvements.

Two practical obstacles have repeatedly limited current machine-learned OF-DFT approaches.
First, learning an energy surface from ground-state labels alone is challenging: during density optimization, the model is queried on densities far from the ground state, yet these off-equilibrium regions are typically weakly constrained by supervised training.
Indeed, generating additional labeled densities away from the minimum turned out to be instrumental to learn a fully convergent functional\cite{remme2025stable}, making the associated effort worthwhile for fully supervised approaches.
But still, these training densities are non-interacting v-representable, which is not guaranteed during optimization. 

Moreover, training data are often generated in representations used in KS-DFT (typically a sum of products of atomic basis functions) and subsequently mapped to a different representation that is used in scalable OF-DFT (typically\cite{M-OFDFT,remme2025stable} a linear combination of atomic basis functions (LCAB), see section \ref{sec:notation}), introducing an additional mismatch.
Second, realizing improved scaling and wall-clock speedups requires that the full pipeline, including density representations and any reparametrizations (such as the $O(N^3)$ Löwdin symmetric orthonormalization\cite{M-OFDFT}) used for stable optimization, remains efficient for large systems.

\begin{figure}[t]
  \includegraphics[width=0.48\textwidth]{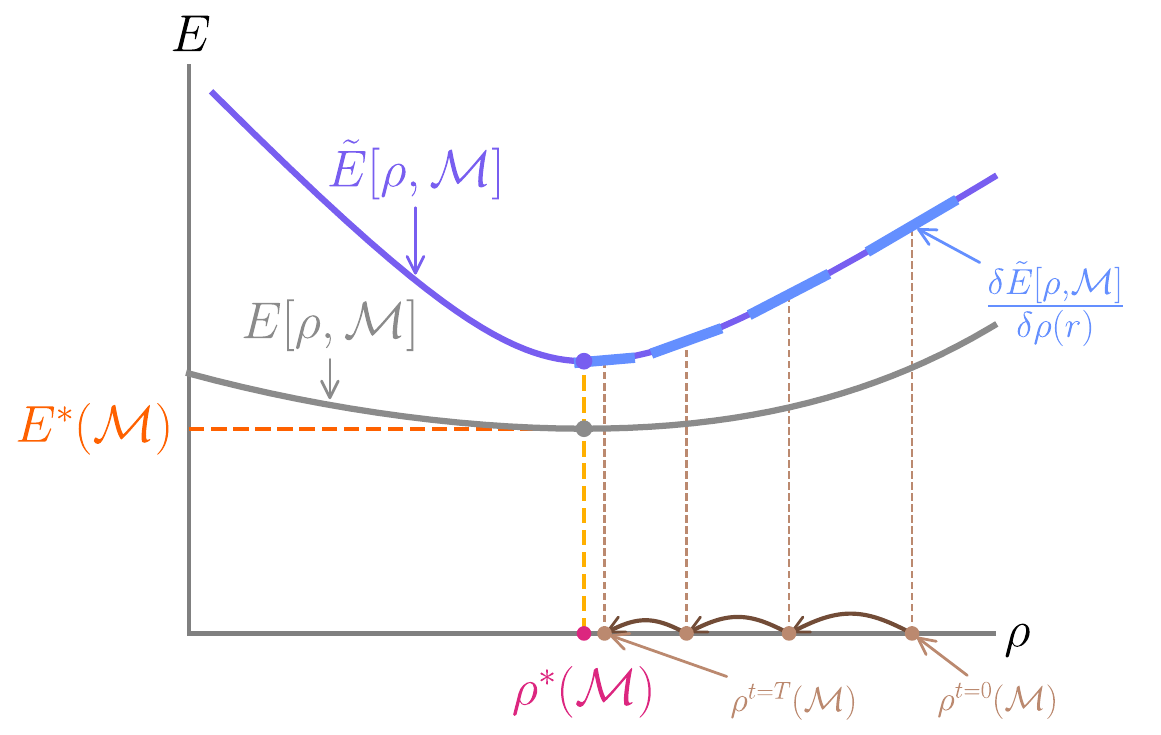}
  \caption[Surrogate Energy Landscape]{\textbf{Surrogate Energy Landscape.} A surrogate energy functional ($\tilde{E}$, violet) can be used in place of the true, physical functional ($E$, gray) in density optimization (brown), resulting in the correct ground-state density $\rho^*(\mathcal{M})$.}
  \label{fig:surrogate_energy_surface}
\end{figure}

We address both issues by introducing \textit{surrogate functionals}.
Rather than requiring a learned functional to be globally faithful to a physical reference, we define it through its role in a fixed density optimization procedure: a surrogate functional is one that yields the true ground-state density coefficients (at least approximately) when minimized by the chosen optimizer from a prescribed initialization (see Figure \ref{fig:surrogate_energy_surface} for an illustration).

Our training approach builds on the observation that even when energy labels are only available at the ground state, we can still impose additional conditions on energies and gradients at arbitrary densities that facilitate successful optimization.
We introduce a \textit{surrogate loss function} that can be evaluated on off-equilibrium densities using only labels at the ground state, and we combine it with an adaptive training scheme that performs density optimization already during training via a caching mechanism.
Together, these ideas focus model capacity and supervision on those parts of density space that matter most for OF-DFT in practice: the optimization trajectories connecting an initial guess to the ground state.

In summary, we
\begin{itemize}
  \item  introduce the concept of \textit{surrogate functionals} for machine-learned OF-DFT and frame their training through the lens of energy-based models\cite{lecun2006tutorial}.
  \item  present a training approach based on a surrogate loss function and train-time density optimization that targets stable OF density optimization while maintaining favorable scaling in the overall pipeline.
  \item  demonstrate that surrogate functionals enable convergent density optimization without the $O(N^3)$ orthonormalization step required by previous work\cite{M-OFDFT,remme2025stable}, improving runtime scaling.
\end{itemize}

\section{Related Work}

\paragraph{Energy-based models.}
Energy-based models (EBMs)\cite{lecun2006tutorial} frame prediction as energy minimization: a neural network defines a scalar energy over input--output pairs, and inference selects the output with the lowest energy.
This paradigm has been applied to structured prediction\cite{belanger2016structured}, generation\cite{du2019implicit}, and classification\cite{grathwohl2019classifier}.
Surrogate functionals can be seen as a special case of EBMs where inference corresponds to density optimization.
A related line of work is score-based generative modeling\cite{song2019generative}; our gradient-based surrogate loss shares a similar spirit, as it directly supervises the gradient field of the energy functional, but, beyond targeting a single ground state rather than a distribution, our approach imposes constraints on the optimization dynamics induced by the gradient field, rather than requiring it to represent a globally consistent score function.

\paragraph{Machine-learned OF-DFT.}
In recent decades, semi-empirical kinetic energy approximations for OF-DFT have made great progress\cite{thakkar1992comparison,wang1992nonlocal,huang2010nonlocal,constantin2011semiclassical,ke2013angular-momentum,luo2018simple,shao2021revised,xu2022nonlocal,mi2023orbital}, far exceeding the classical Thomas--Fermi model\cite{thomas1927calculation, fermi1928statistische} and von Weizs\"acker models\cite{von1935theorie}, working especially well when combined with pseudopotentials on metallic systems\cite{shao2021revised,xu2022nonlocal}.
However, they still struggle on molecular systems, where recent machine learning approaches have started to address this gap:
Pioneering work in machine-learned OF-DFT provided proof of concept on 1D toy systems\cite{snyder2013orbital,meyer2020machine}, 
later contributions generalized to small molecules\cite{yao2016kinetic,seino2018semi-local,golub2019kinetic,fujinami2020orbital-free,kineticnet}. M-OFDFT\cite{M-OFDFT} represented a step change in scope, demonstrating high-accuracy machine-learned kinetic energy density functionals for diverse molecular systems, but still failed to learn a convergent functional.
STRUCTURES25\cite{remme2025stable} further improved accuracy, but more importantly was the first to achieve reliable convergence in density optimization across the QM9\cite{ramakrishnan2014quantum} and QMugs\cite{isert2022qmugs} datasets.
All of these approaches train supervised models to approximate a physical energy functional; in contrast, the present work defines success through the outcome of density optimization, relaxing the requirement of global fidelity to the physical functional.

\section{Methods}
\subsection{Notation: OF-DFT on a linear combination of atomic orbitals}\label{sec:notation}
Following previous work\cite{M-OFDFT, remme2025stable} on machine-learned density functionals, we employ a linear combination of atomic basis functions (LCAB) Ansatz\cite{grisafi2018transferable, vergara2023efficient} to express the electron density: 
\begin{align}
  \rho(\vecr) = \sum_{\mu}p_{\mu}\omega_{\mu}(\vecr)\,, \label{eq:LCAB}
\end{align}
where the coefficient vector $\{p_\mu\}$ weighs the atom-centered Gaussian basis functions $\omega_{\mu}$.
In this density representation, the density functional $E[\rho]$ becomes a function $E(\vp)$ and the baseline application of OF-DFT of finding the electronic ground state via density optimization amounts to finding
\begin{align}
  \vp^* = \argmin_{\vp} E(\vp)\,. \label{eq:density_optimization}
\end{align}
This representation is parameter efficient (in particular compared to grid-based representations), simple and allows for fast computation of integrals involving the electron density, particularly in the case of Gaussian basis functions.
However, we mainly choose it here for direct comparison with prior work and note that our work straightforwardly generalizes to other density representations, e.g.~grid-based approaches or numerical basis functions. %

\subsection{Surrogate Functionals}\label{sec:surr_def}
We define a (successfully trained) ``surrogate functional'' only in the context of a density optimization procedure:
\begin{definition}
    A \textbf{density optimization procedure} is a pair formed by an initial estimator that maps molecules $\mathcal{M}$ to initial density coefficients $\vp^{(0)}$, and a (usually iterative) optimizer, which, given an energy functional $E$, maps $\mathcal{M}$ and $\vp^{(0)}$ to final density coefficients $\vp^{(T)}$.
\end{definition}
Thus, the density optimization procedure includes the choice of initial estimate (e.g.~a superposition of atomic densities, SAD), as well as the choice of optimizer (e.g.~gradient descent) and all of its parameters (e.g.~learning rate, momentum).
\begin{definition}
    A \textbf{surrogate functional} for a given density optimization procedure is a functional which, when used in lieu of the true energy functional in density optimization, leads to the true ground-state coefficients $\vp^{(T)}=\vp^*$.
\end{definition}
Hence, in terms of finding the ground state, a surrogate functional can perfectly replace the physical ``ground-truth'' electronic energy functional.

Note that, as defined, a surrogate functional is not required to predict the correct ground-state energy; it only needs to lead to the correct ground-state density under the given optimization procedure, and we choose to focus on this setting in the present work. We discuss the natural extension to \textit{strong surrogate functionals}, which additionally reproduce the ground-state energy, in section~\ref{sec:conclusion}.

So far, we have only defined surrogate functionals to replace the total energy functional. However, surrogate functionals which replace only parts of the energy functional are also conceivable. These would give rise to a surrogate for the total energy after exact expressions for the other contributions are added.
We defer exploration of this avenue to future work, with the principal difficulty being the efficient computation of values and gradients for those portions of the energy functional excluded from the machine learning model.

\subsection{Surrogate Loss Function}\label{sec:surr_losses}
A key question is how to train a surrogate functional. Crucially, surrogate loss functions can be evaluated at \textit{any} density coefficients $\vp$, as long as the ground-state coefficients $\vp^*$ are known---no additional labels away from the ground state are required. Among several conceivable surrogate losses (see section~\ref{sec:conclusion} for a discussion of alternatives), we focus on one that most directly addresses the goal of density optimization: the \textit{gradient-descent-improvement} (GDI) \textit{loss}.

The idea is to require that every gradient descent step moves the density coefficients closer to the true ground state $\vp^*$. Concretely, we demand that the distance to the ground state decrease by at least a factor $0 < \beta < 1$ in every step:
\begin{align}\label{eq:grad_improvement_loss}
    &\mathcal{L}_{\text{GDI}} = \max\big(0, \norm*{\underbrace{\vp - \lambda \grad_\vp \tilde{E}(\vp; \theta)}_{\text{coeffs after step}} - \vp^*} - \beta\norm{\vp - \vp^*}\big) \,.
\end{align}
Here, $\lambda$ is the step size used in gradient descent during density optimization.

A key advantage of this loss is that if successful model training makes it zero for all densities of a test system, then density optimization with gradient descent is guaranteed to converge to the true ground state, and to do so quickly: if the distance of the initial guess to the ground state is $d$, the distance after $n$ steps is at most $d \beta^n$. This provides a principled convergence guarantee that is directly tied to the optimization procedure.

The contraction factor $\beta$ controls the trade-off between the restrictiveness of the loss and its compatibility with the true energy functional. For conservative (i.e.~high) values of $\beta$ close to 1, the loss is easily satisfiable and compatible with a wider range of functionals, but convergence may be slow. For aggressive (i.e.~low) values, the loss demands rapid convergence, which may not be achievable by the true energy functional but can still be realized by a surrogate.

\begin{figure}[t]
  \centering
  \includegraphics[width=0.45\textwidth]{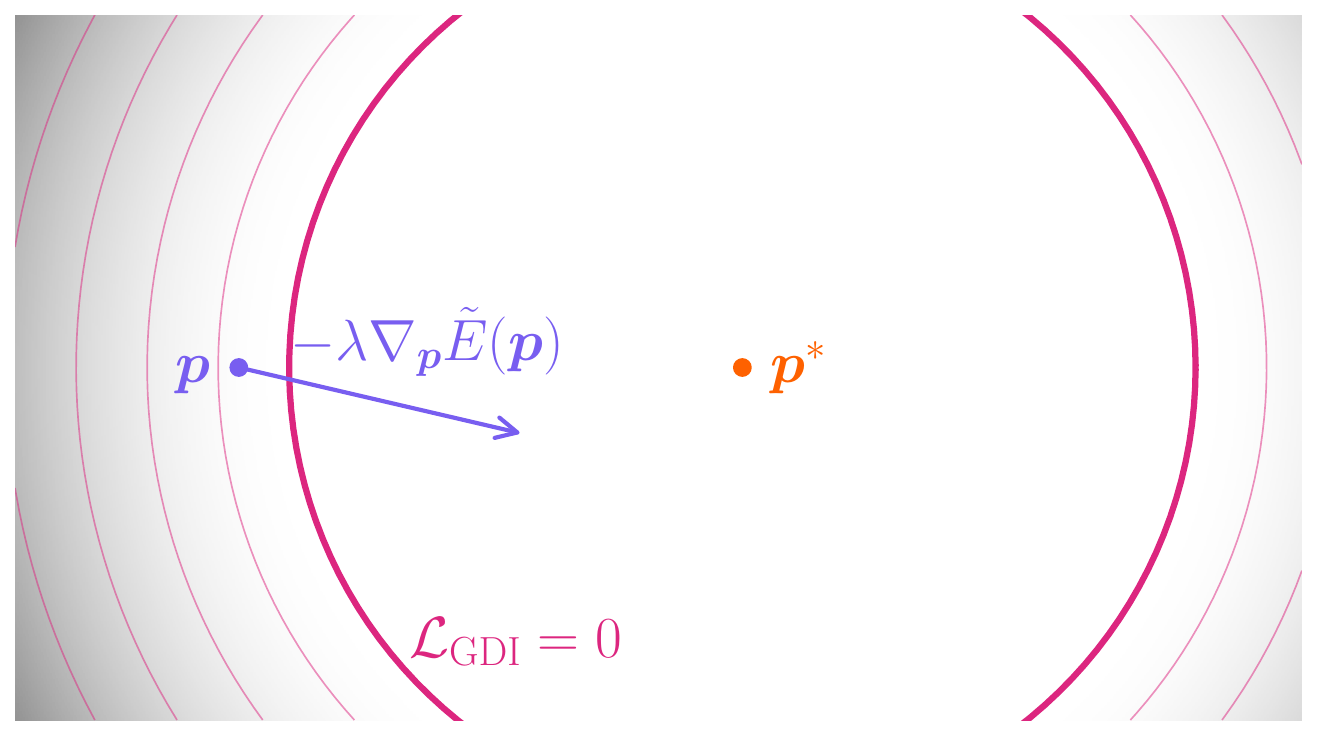}
  \caption[Isocontours of the GDI loss]{\textbf{Isocontours of the GDI loss.} For the loss to be zero, the gradient descent update step (purple) must reduce the distance to the ground state by a factor of $\beta=0.9$, i.e.~lead to coefficients inside the magenta circle.}\label{fig:grad_improvement_isocontours}
\end{figure}

\begin{figure}[t]
  \centering
  \includegraphics[width=0.45\textwidth]{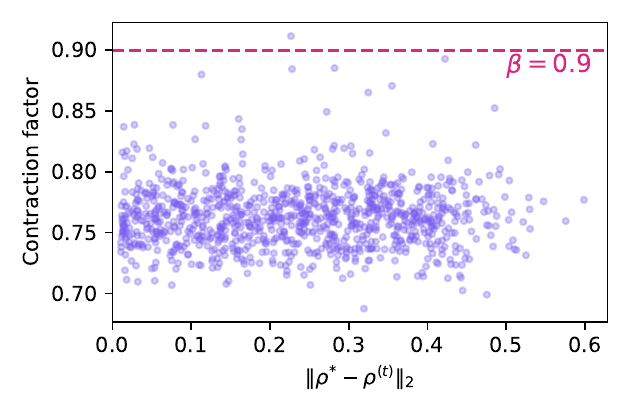}
  \caption[Contraction factors on labeled data]{\textbf{Contraction factors on labeled data.} Each point corresponds to a sample from the perturbed QM9 dataset~\cite{remme2025stable} after Löwdin symmetric orthonormalization. The horizontal axis shows the $L_2$ distance of the sample's density coefficients to the ground state; the vertical axis shows the contraction factor, i.e.~the ratio $\|\vp - \lambda\grad_\vp E(\vp) - \vp^*\| / \|\vp - \vp^*\|$, for a single gradient descent step with the ground-truth gradient and a step size of $\lambda = 0.05$. Nearly all points lie below $\beta = 0.9$, indicating that this choice of contraction factor is broadly compatible with the physical energy functional on this data.}\label{fig:contraction_factor}
\end{figure}

\subsection{Train-time density optimization}
An ideal machine-learned energy functional would perfectly generalize over all of chemical space and the space of all possible input density coefficients $\vp$. In practice, however, a trade-off between the set of input densities which the model generalizes over and the accuracy of the model will likely be necessary.

Here, we are in a fortunate situation: the surrogate loss function introduced above (see section \ref{sec:surr_losses}) is applicable to \textit{any} set of coefficients $\vp \in \mathbb{R}^n$. The question is therefore not \emph{whether} we can evaluate the loss, but \emph{where} in coefficient space we should place training signal. 
This is in contrast to supervised training, where generating labels for arbitrary densities would in principle require use of inverse DFT, which is expensive and numerically non-trivial. %

To achieve successful density optimization, the minimal set of densities the model has to work well on is the density optimization path between the initial guess and the ground-state coefficients, or, to be robust, a neighborhood of this path. With a good initial guess (which can be easily learned~\cite{M-OFDFT}), sampling coefficients isotropically around the ground state can in principle cover such paths. In practice, however, we have found that models tend to exploit ``loopholes'' under such static sampling (see Appendix \ref{app:loopholes} for a detailed discussion). This motivates an adaptive strategy which performs density optimization on the fly during training and thereby focuses learning on the actual trajectories that the optimizer follows.

A direct way to achieve this would be to unroll multiple density optimization steps for each training sample and apply the surrogate loss along the resulting trajectory. While conceptually clean, this approach is expensive: the per-iteration cost grows with the number of unrolled steps, and GPU memory limits the feasible trajectory length. Instead, we adapt the idea of persistent contrastive divergence (PCD)\cite{tieleman2008training} to our setting, maintaining per-molecule persistent coefficients that are advanced by one optimization step whenever the molecule is seen.

Concretely, each molecule $\mathcal{M}$ stores a cached coefficient vector $\vp^{(t)}$ (one per molecule). When a batch is loaded, we replace the coefficients of any molecule present in the cache with its cached value, compute energies and gradients, apply the surrogate loss, and then take a single density optimization step using the same optimizer as during inference. The updated coefficients are written back to the cache. To prevent the cache from drifting too far and to ensure that early trajectory regions remain represented in training, we reset cached coefficients with probability $q_{\text{reset}}=0.01$ to a fresh perturbation around the ground state. The perturbation direction is sampled uniformly, while the radius is drawn from a Gaussian distribution centered at a typical distance between the initial guess and the ground state, see Appendix \ref{app:perturbation}. This yields long effective trajectories over the course of training without expensive unrolling, and it concentrates learning on precisely the densities encountered during optimization. Figure~\ref{fig:train_time_denop} summarizes the procedure and pseudocode is provided in Appendix~\ref{app:train_time_denop}.

\begin{figure*}[t]
  \centering
  \includegraphics[width=0.8\textwidth]{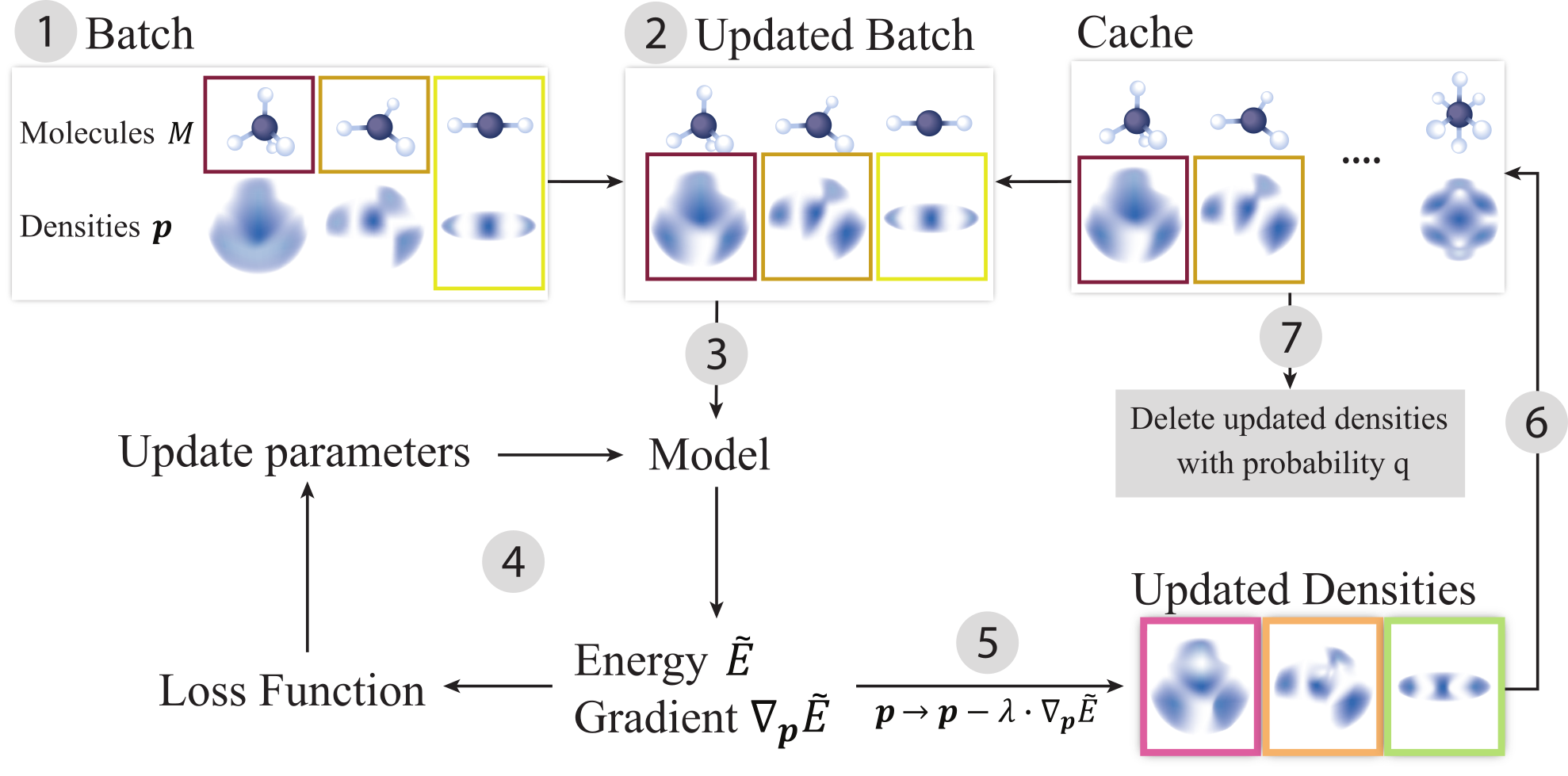}
  \caption[Train-time on-the-fly density optimization via caching]{\textbf{Train-time on-the-fly density optimization via caching.} To train on densities akin to those encountered at inference time during density optimization, we adapt persistent contrastive divergence\cite{tieleman2008training}:  A training batch is loaded (1), updated with densities of all molecules in the batch which are present in the cache (2), passed through the model (3) yielding energies and gradients which are both used 
  in the model update loop (4)
  as well as utilized in a density optimization step (5). The updated densities are written to the cache (6), before finally each molecule of the present batch is discarded from the cache with probability $q_{\text{reset}}$ (7).}\label{fig:train_time_denop}
\end{figure*}

\subsection{Model Architecture}
For comparison with the current state of the art in molecular OF-DFT, we mostly follow the architectural choices of prior work\cite{remme2025stable}: We employ a modified Graphormer architecture\cite{ying2021transformers}, enhanced with tensorial message passing.\cite{lippmann2025beyond}
Notable changes are to the initial dimension-wise rescaling (see Ref.~\cite{M-OFDFT} for details) of input coefficients: For training of surrogate models, we replace the scaling according to a trade-off between coefficient variance and gradient norm (optimized for supervised training of the variational model) by a simple scalar factor of 10, as we have found that networks otherwise sometimes completely ignore some input coefficients with small prefactors

Furthermore, we replace the atomic reference module, which in its original form\cite{M-OFDFT} is a linear fit to the energy which is being added to the neural network output after the final layer, with a simple parabola around the coefficients of the superposition of atomic densities (dSAD\cite{remme2025stable}) $\bar{\vp}$:
\begin{align}
  \vp \mapsto a \cdot \norm{\vp - \bar{\vp}}^2\,.
\end{align}
We found that a prefactor of $a=0.1$ works well.

\section{Results}
\begin{figure}[h]
    \centering
    \hfill
    \begin{subfigure}{0.2\textwidth}
        \centering
        \includegraphics[angle=90, width=\linewidth]{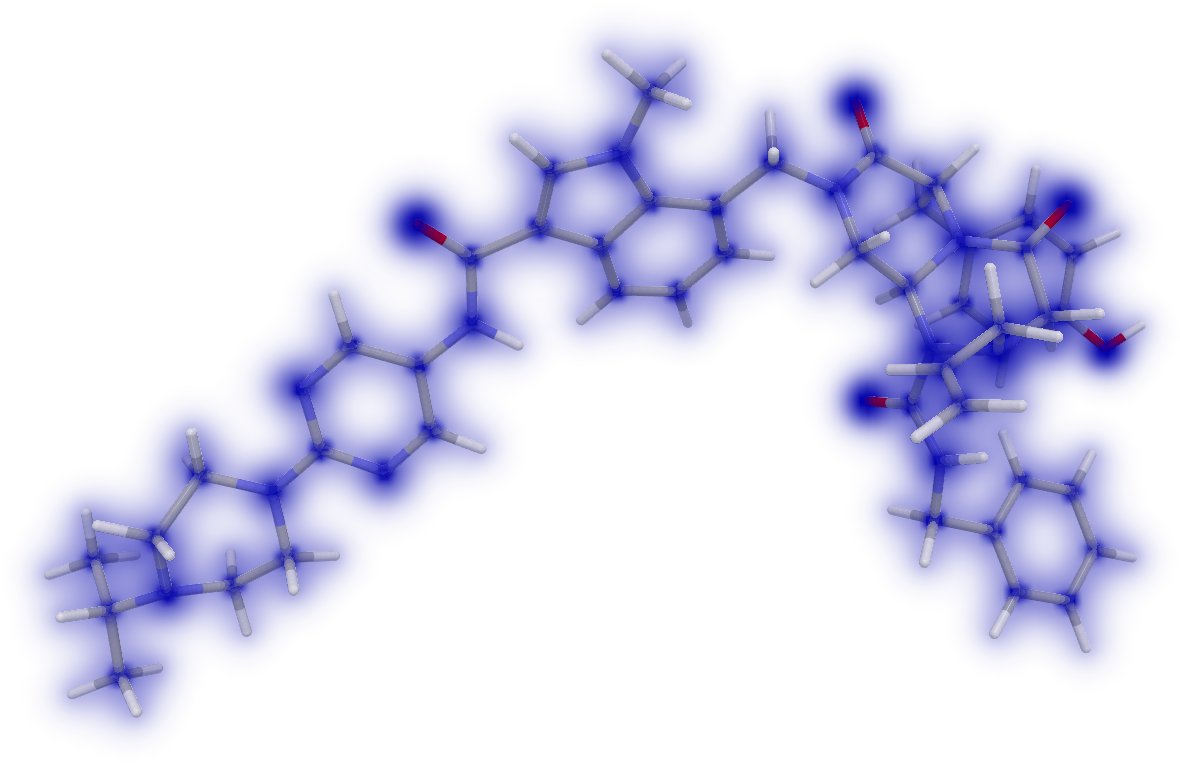}
        \caption{}
        \label{fig:a}
    \end{subfigure}
    \hfill
    \begin{subfigure}{0.2\textwidth}
        \centering
        \includegraphics[angle=90, width=\linewidth]{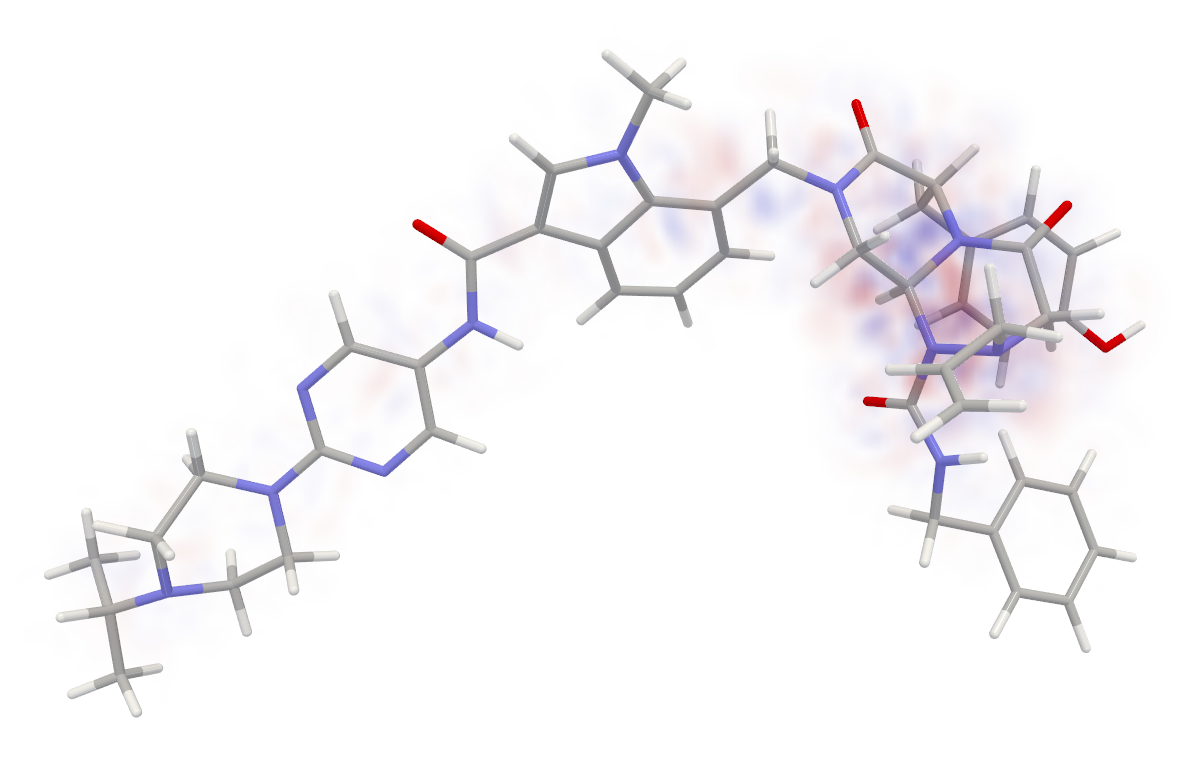}
        \caption{}
        \label{fig:b}
    \end{subfigure}
    \hfill
    \caption{\textbf{Electron density prediction and error for a QMugs molecule.} The molecule, $\text{H}_{53} \text{C}_{46} \text{N}_{11} \text{O}_{5}$, has near-average molecular weight within the QMugs test set. (a) Converged ground-state density. (b) Difference to ground-truth density, scaled by a factor of 20 for visibility.}
    \label{fig:QMUGS_example_density_and_error}
\end{figure}
\begin{table}[t]
  \caption{\textbf{Density optimization accuracy and runtime on QM9\cite{ramakrishnan2014quantum} and QMugs\cite{isert2022qmugs}.} Comparison to state-of-the-art machine-learned OF-DFT approaches (M-OFDFT\cite{M-OFDFT} and STRUCTURES25\cite{remme2025stable}) and our surrogate functional, with and without the $O(N^3)$ Löwdin symmetric orthonormalization step. We report the $L_2$ density error $\|\Delta \rho\|_2$ and the average runtime per molecule for density optimization.}
  \label{tab:results}
  \centering
  \footnotesize
  \begin{ruledtabular}
  \begin{tabular}{llccr}
    Dataset & Functional & \begin{tabular}[c]{@{}c@{}}Avoids $O(N^3)$\\orthonormalization\end{tabular} & $\| \Delta \rho \|_2$ ($\text{10}^{-\text{2}}$) & Runtime (s)\\
    \hline
    QM9 & M-OFDFT~\cite{M-OFDFT} & $\times$ & 2.7 & 183\hspace{3ex}~\\
    & STRUCTURES25~\cite{remme2025stable} & $\times$ & $1.40 \pm 0.02$ & 13\hspace{3ex}~\\ %
    & Ours (natrep) & $\times$ & 1.2 & 7\hspace{3ex}~\\
    & Ours (no-natrep) & $\checkmark$ & 1.2 & 8\hspace{3ex}~\\
    \hline
    QMugs & M-OFDFT~\cite{M-OFDFT} & $\times$ & 7.0 & 319\hspace{3ex}~\\
    & STRUCTURES25~\cite{remme2025stable} & $\times$ & $6.8 \pm 0.2$ & 40\hspace{3ex}~\\ %
    & Ours (natrep) & $\times$ & 8.2 & 20\hspace{3ex}~\\
    & Ours (no-natrep) & $\checkmark$ & 12.0 & 21\hspace{3ex}~\\
  \end{tabular}
  \end{ruledtabular}
\end{table}
\subsection{Density optimization accuracy and runtime}
Table~\ref{tab:results} compares density optimization performance to state-of-the-art machine-learned OF-DFT approaches.
Previous work (STRUCTURES25 and M-OFDFT) relies on an $O(N^3)$ reparametrization step of the density coefficients to stabilize optimization, which dominates cost as the system size grows.

In contrast, our surrogate models achieve convergent density optimization both with Löwdin symmetric orthonormalization (\textit{natrep}), but also directly in coefficient space (\textit{no-natrep}), improving the asymptotic complexity of the entire pipeline and realizing improved scaling with system size compared to Kohn-Sham DFT. 
Regarding accuracy, on QM9 we obtain essentially identical density errors in both settings ($\|\Delta\rho\|_2 = 1.2\times 10^{-2}$), completely eliminating the need for orthonormalization for these small molecules and improving upon prior work.
We follow the experimental setup of prior work\cite{M-OFDFT,remme2025stable} in evaluating extrapolation to larger systems using molecules from the QMugs\cite{isert2022qmugs} dataset. Figure \ref{fig:QMUGS_example_density_and_error} shows the converged electron density and its deviation from the reference for an exemplary test molecule. On QMugs, removing symmetric Löwdin orthonormalization incurs a moderate degradation ($0.082 \to 0.12$), resulting in errors that are slightly higher than the best prior results but remain in the same order of magnitude.
Regarding run-time, surrogate models improve throughput on both datasets, in large part due to a smaller number of steps being required to reach convergence, see Table \ref{tab:results}.

\subsection{Choosing surrogate loss hyperparameters}
The GDI loss (Equation~\ref{eq:grad_improvement_loss}) introduces two hyperparameters: the contraction factor $\beta$ and the step size $\lambda$.

For $\beta$, we follow a data-driven strategy. Because the GDI loss is satisfiable by the true energy functional whenever the physical gradient yields a contraction factor below $\beta$, we can check compatibility on labeled data. Figure~\ref{fig:contraction_factor} shows the empirical contraction factor for each sample in the perturbed QM9 dataset~\cite{remme2025stable}, computed from ground-truth gradients at a fixed step size. Nearly all samples achieve a contraction factor below $0.9$, which motivates our choice of $\beta = 0.9$. This value is conservative enough to be broadly compatible with the physical functional, while still guaranteeing fast convergence.

For the step size, we use $\lambda = 0.1$ for density optimization, which we found to work well empirically. Note that $\lambda$ and the energy scale are coupled: multiplying $\lambda$ by a constant $c$ has the same effect on the gradient descent update $\vp - \lambda \grad_\vp \tilde{E}$ as rescaling the energy functional by $c$, i.e.~$\tilde{E} \to c\,\tilde{E}$. Hence, the step size does not constrain the expressiveness of the surrogate functional; it merely fixes a scale for the learned energy surface.

\section{Discussion and Conclusion}\label{sec:conclusion} 
In the following, we relate the current contribution to direct prediction of ground-state densities and discuss alternative surrogate losses and ``strong'' surrogate functionals which also predict physical energies.

\paragraph{Surrogates vs.\ direct prediction.}
In principle, one could learn a trivial surrogate: an isotropic parabola centered at the ground state, $\tilde{E}(\vp) = a\|\vp - \vp^*\|^2$, for which gradient descent provably converges to $\vp^*$.
However, learning such a parabola is essentially equivalent to direct ground-state prediction: the center $\vp^*$ can be recovered from the energy and gradient at any single point (the direction to $\vp^*$ from the gradient, the distance from the energy), so the model would need to implicitly encode the exact ground state in its predictions at every input density.
We hypothesize that the advantage of surrogate functionals lies precisely in the freedom to learn energy surfaces that are \textit{not} simple parabolas but rather reflect the structure of the underlying physics.
Such functionals may be easier to learn and generalize better, because the model can leverage physically meaningful relationships between input density and energy, for instance by trading off kinetic and potential contributions given a particular input density, rather than encoding the ground-state coefficients directly.
In other words, it may be easier for a model to predict an energy landscape whose gradients \textit{roughly} point towards the ground state than to implicitly encode the exact minimizer at every point in coefficient space.

\paragraph{Alternative surrogate losses.}
While we focus on the GDI loss, many other surrogate loss functions are conceivable. For instance, a \textit{lower-bound loss} motivated by the variational principle could require that the learned functional assigns any density an energy no lower than that of the ground state, e.g.\ $\mathcal{L}_{\text{lb}} = \max(0, \tilde{E}(\vp^*; \boldsymbol{\theta}) - \tilde{E}(\vp; \boldsymbol{\theta}))$. This is fully compatible with the true energy functional but does not prevent local minima and is therefore insufficient on its own. A \textit{gradient-to-ground-state loss} could require the negative gradient to point approximately towards $\vp^*$, penalizing low cosine similarity between $\grad_\vp \tilde{E}$ and $\vp - \vp^*$; see appendix~\ref{app:grad_to_gs_figure} for an illustration. This constrains gradient directions but not norms, which can be addressed by an additional \textit{gradient-norm-range loss} that keeps gradient magnitudes within specified bounds. Among these alternatives, the GDI loss most directly addresses the goal of density optimization by providing explicit convergence guarantees tied to the optimizer, and led to the best results in our experiments.

\paragraph{Strong surrogate functionals.}
As defined in this work, a surrogate functional is only required to yield the correct ground-state density, not the correct ground-state energy. A natural extension is to additionally require that the functional assigns the true ground-state energy $E^*(\mathcal{M})$ to the ground-state coefficients $\vp^*$. Such a \textit{strong surrogate functional} would be a full replacement for the physical energy functional in the most common applications of OF-DFT. Achieving this likely requires combining surrogate losses with supervised energy objectives, for example by adding a standard regression loss on the predicted energy at the ground state. We leave the investigation of strong surrogates to future work.
Possible future work also includes scaling to larger datasets such as OMol25\cite{levine2025open}, where the ability to train on ground-state densities alone is particularly advantageous; and extending the GDI objective to more sophisticated optimizers (e.g.~momentum methods or line search) to further improve convergence behavior.

In summary, we introduce surrogate functionals for machine-learned OF-DFT, considering the problem through the lens of energy-based modeling and defining success by the outcome of a fixed density-optimization procedure. This perspective leads to a surrogate loss function that can be evaluated on arbitrary densities using only ground-state labels, along with a train-time density-optimization strategy that concentrates learning on the densities actually visited during optimization. 
Empirically, surrogate training yields reliable density optimization while reducing the computational complexity of the overall pipeline.

\begin{acknowledgments}
This work is supported by the German Research Foundation (DFG) under Germany's Excellence Strategy EXC-2181/1-390900948 (the Heidelberg STRUCTURES Excellence Cluster) as well as by the Carl-Zeiss-Stiftung via its Wildcard program.
The authors acknowledge support by the state of Baden-W\"urttemberg through bwHPC and the German Research Foundation (DFG) through grant INST 35/1597-1 FUGG.
\end{acknowledgments}

\section*{Author Contributions}
Roman Remme: Conceptualization (lead); Methodology (lead); Software (lead); Writing – original draft (lead); Writing – review \& editing (equal). Fred A. Hamprecht: Conceptualization (supporting); Supervision (lead); Writing – review \& editing (equal); Funding acquisition (lead)

\appendix

\section{Train-time density optimization details}
\label{app:train_time_denop}

This appendix collects implementation details for the train-time density optimization procedure (Figure~\ref{fig:train_time_denop}), including sampling, caching, and a pseudocode listing.

\subsection{Why static sampling can fail (loopholes)}\label{app:loopholes}
Static sampling around the ground state can permit spurious solutions where the model ``solves'' only easy directions. For instance, coefficients describing core densities can be predicted accurately without capturing chemically relevant bonding features. In such a case, cosine-similarity based losses can be satisfied for most sampled points while optimization still fails along the difficult directions. This failure mode also motivated the adaptive, trajectory-based sampling used during training.

\subsection{Initialization around the ground state}\label{app:perturbation}
Whenever cached coefficients are reset, we sample a fresh perturbation around the ground-state coefficients $\vp^*$.
We draw a random direction $\vu$ uniformly on the unit sphere in $\mathbb{R}^n$ and a radius $r$ from a Gaussian distribution centered at a distance of $0.05$ with standard deviation $0.05$.
The reset coefficients are then
\begin{equation}
  \vp^{(0)} = \vp^* + r\,\vu\,. \label{eq:perturbation}
\end{equation}
This construction ensures isotropic perturbations while controlling the typical distance from $\vp^*$ via the mean of the Gaussian.

\subsection{Cache structure and updates}
We maintain one cached coefficient vector per molecule. At training time, if a molecule appears in a batch and is present in the cache, its coefficients are replaced by the cached value before computing losses. After a single density optimization step, the updated coefficients are written back to the cache. With probability $\qreset$, we discard the cached coefficients and reinitialize them as described above.

In our implementation, the cache is stored on the GPU, which is feasible for the datasets used here (a few GB). For substantially larger datasets, the cache could be moved to CPU memory or disk at the cost of additional data transfer overhead.

\subsection{Dataset structure and number of updates}
During training, we replace the density coefficients from the labeled dataset with the cached coefficients. Each molecule appears 21 times in the training data\cite{remme2025stable}, and therefore each epoch performs up to 21 train-time density optimization steps per molecule, possibly fewer if the molecule is drawn multiple times within the same batch. This may be preferable over a dataset where every molecule appears exactly once per epoch, as in the latter case all molecules would progress synchronously in train-time density optimization, while the random ordering of the dataset with duplicates leads to more varied progress in train-time density optimization.

\subsection{Pseudocode}\label{app:pseudocode}

\begin{algorithm}[h]
Initialize empty cache $C$.\\
\For{each training step}{
    Load batch $B$ of molecules and ground-state coefficients.
    \For{$(\mathcal{M}, \vp) \in B$}{
        \If{$\mathcal{M} \in C$}{ $\vp \leftarrow C[\mathcal{M}]$ }
        \Else{ $\vp \leftarrow \text{perturb}(\vp)$ (see Eq. \ref{eq:perturbation})}
    }
    Evaluate model; compute energies and gradients for $B$.\\
    Evaluate surrogate loss and update model parameters. \\
    \For{$(\mathcal{M}, \vp) \in B$}{
        Take density-optimization step: $\vp \leftarrow \vp - \lambda \grad_\vp \tilde{E}(\vp,\mathcal{M}; \theta)$\\
        Store new coefficients in cache: $C[\mathcal{M}] \leftarrow \vp$\\
        \If{random() $< q$}{ Delete $\mathcal{M}$ from $C$ }
    }
}
\caption{Train-time Density Optimization with Caching}
\end{algorithm}

\section{Gradient-to-ground-state loss: energy surface illustration}
\label{app:grad_to_gs_figure}
\begin{figure}[t]
    \centering
    \includegraphics[width=0.45\textwidth]{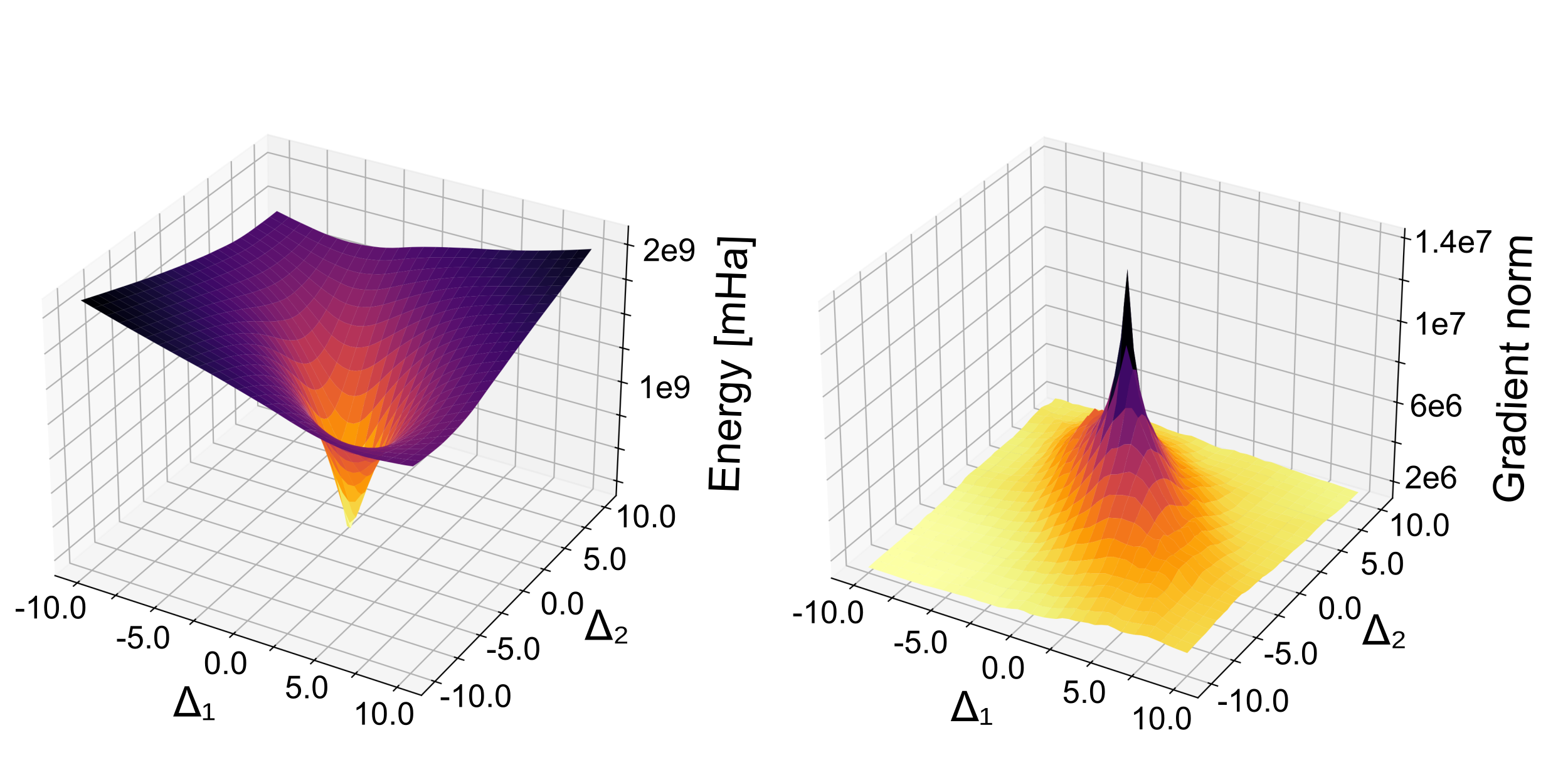}
    \caption[2D Slice of energy and gradient-norm surface trained with gradient-to-ground-state loss alone]{\textbf{2D Slice of energy and gradient-norm surface trained with gradient-to-ground-state loss alone.} These plots show the energy surface (left) and gradient norm (right) of a preliminary model trained only with a gradient-to-ground-state loss on a 2D slice of the input space, spanned by the directions from the ground state to the two penultimate SCF iterations. While a minimum in the correct position is apparent, the gradient in its vicinity is not well behaved, hindering density optimization. This is why the main text instead reports results from the GDI loss introduced in section \ref{sec:surr_losses}}\label{fig:grad_to_gs_energy_surfaces}
\end{figure}
Figure~\ref{fig:grad_to_gs_energy_surfaces} shows the energy surface and gradient norm of a preliminary model trained with the gradient-to-ground-state loss alone (see section~\ref{sec:conclusion} for a description of this loss), on a 2D slice of the input space.
This illustrates a possible disadvantage of gradient-direction-based losses: they do not restrict the norm of the predicted gradient, only its direction via the cosine similarity.
When used on their own, this can lead to vastly varying gradient scales which may be problematic during density optimization.

\section*{References}
\bibliography{bibliography}%

\end{document}